# Improving Automatic Emotion Recognition from speech using Rhythm and Temporal feature

Mayank Bhargava[a],a*, Tim Polzehl[b]

[a] School of Electrical Sciences, Indian Institute of Technology Bhubaneswar, Bhubaneswar 751013, India
[b] Quality and Usability Lab, T labs/ TU Berlin, Germany

## Abstract

This paper is devoted to improve automatic emotion recognition from speech by incorporating rhythm and temporal features. Research on automatic emotion recognition so far has mostly been based on applying features like MFCC's, pitch and energy/intensity. The idea focuses on borrowing rhythm features from linguistic and phonetic analysis and applying them to the speech signal on the basis of acoustic knowledge only. In addition to this we exploit a set of temporal and loudness features. A segmentation unit is employed in starting to separate the voiced/unvoiced and silence parts and features are explored on different segments. Thereafter different classifiers are used for classification. After selecting the top features using an IGR filter we are able to achieve a recognition rate of 80.60 % on the Berlin Emotion Database for the speaker dependent framework.

## 1. Introduction

Automatic emotion recognition from speech signal has become a major research topic in the field of Human computer Interaction (HCI) in the recent times due to its many potential applications. It is being applied to growing number of areas such as humanoid robots, car industry, call centers, mobile communication, computer tutorial applications etc. [1]. In this paper we focus on emotion recognition from acoustic properties of speech.

The most commonly used acoustic features in literature are related to MFCC's and prosody features like pitch, intensity and speaking rate. In addition to these features we exploit a set of loudness features which are extracted by using bark filter bank as described by Zwicker[2]. Rhythm based features which have widely been used in music emotion recognition and in automatic speech assessment applications based on the linguistic properties of speech are exploited in this paper on the basis of acoustic knowledge only. Speech rhythm can be understood as a measure describing the regularity of occurrence of certain language elements in speech that are perceptually similar, e.g., sequences of stressed syllables. We investigate different metrics of speech rhythm with the aim to study their relevance for the characterization of emotion from speech. We also focus on the temporal features which are related to the fluency of speech such as the ratios of pause to voiced parts, pause to unvoiced parts, ratio of voiced to unvoiced parts etc. All these features are explained in detail in Section 3. The overall system is shown in Fig. 1.

In our initial step of the system we implement a segmentation algorithm to separate voiced, unvoiced and pauses in the speech signal. This is a very important part as features are extracted on voiced and unvoiced segments separately and temporal features are mainly based on proper segmentation. Next is the Feature extraction unit where we extract a set of 487 features. In the feature selection part we use Information Gain Ratio (IGR) filter for selecting the best features for classification. For classification we use two different classifiers Artificial Neural Network (ANN) and Support Vector Machine (SVM). We do classification on three different levels: firstly we do the 7 class classification, next we break the 7 class into binary classification targets and then do classification and lastly we break the 7 class in High Arousal and Low arousal sets and then do classification.

In this paper, the Berlin Emotional database (EMO-DB) [4] is used as a database for experiments, which contains 535 utterances of 10 actors (5 male, 5 female) simulating 7 emotional states. The emotions are anger (Wut), boredom (Langeweile), disgust (Ekel), fear (Angust), happiness (Freude), sadness (Trauer) and a neutral emotional state.

This paper is organized as follows: Section 2 reviews the previous research .Section 3 describes our Segmentation unit. Section 4 focuses on Feature extraction and Section 5 on Feature Selection. Section 6 shows the classification results and Section 7 contains a conclusion.

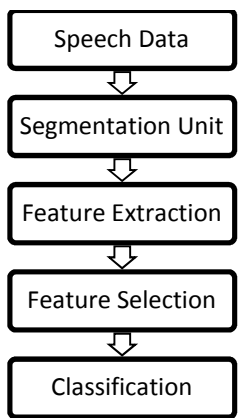

Fig. 1. Speech Emotion Recognition System

## 2. Previous Research

A lot of research has been done in the field of automated speech emotion recognition. In terms of rhythm features Ramus et al. [9] proposed the three metrics based on speech unit internals, including V, the standard deviation of vocalic internals (Vs); C, the standard deviation of consonantal internals (Cs); and %V, the relative duration of vocalic intervals within the total utterance. Low et al. [10] proposed the pairwise variability index (PVI) metrics, to capture the sequential nature of rhythmic contrasts. Dellwo [17] utilized VarcoC, which is the standard deviation of consonantal interval durations (Cs) divided by the mean consonantal durations multiplied by 100. These features have widely been used in linguistic based applications [8, 12]. Lei and Claus used rhythm features to automatically assess non-native speech [12]. These features have widely been used in music emotion recognition applications [3]. In [15] study is done on how temporal characteristics of human speech (e.g. segmental or prosodic timing patterns, speech rhythmic characteristics and durational patterns of voicing) contribute to speaker individuality. In [16] temporal features are used for emotion recognition in spoken Finnish.

## 3. Segmentation Unit

First step towards segmentation of speech signal into voiced, unvoiced and silence is Voice Activity Detection (VAD). We use the algorithm proposed by Rabiner and Sambur for doing the VAD [5]. Firstly we calculate Zero Crossing Rate (ZCR) and Short Time Energy for each of the 30 ms long, 50 % overlapping frames obtained after applying a rectangular window function on the speech signal. Then simple energy based threshold detection is done to estimate the active part. Next a smoothing algorithm is applied. Then we use the ZCR to extend the endpoints of an active area. This concludes the VAD.

The autocorrelation function of a segment of voiced speech should show higher values that the one of an unvoiced speech sound. Voiced segments generally also have low ZCR Value. So we keep a low voicing and ZCR threshold to avoid losing many voiced segments. Then we check for other segments in the active part which are not classified as voiced and have ZCR value greater than a set threshold and classify them as unvoiced. The rest of segments in active region are classified as silence/pauses. Figure 2 displays the output of our segmentation unit. For example from time interval 1.3 to 1.6 ms approx. the value of ZeroCrossing (dark blue line) is well above the ZCR threshold (magenta line) and also energy (light blue line) is present in that interval so it is classified as unvoiced. Similarly from time interval 0.2 to 0.4 second energy (light blue line) and harmonicity (bold dark green line) are high whereas ZeroCrossing (dark blue line) is less than threshold so it is classified as voiced. In between interval 0.4 to 0.6 there is no energy (light blue line) so it is classified as silence/ pause.

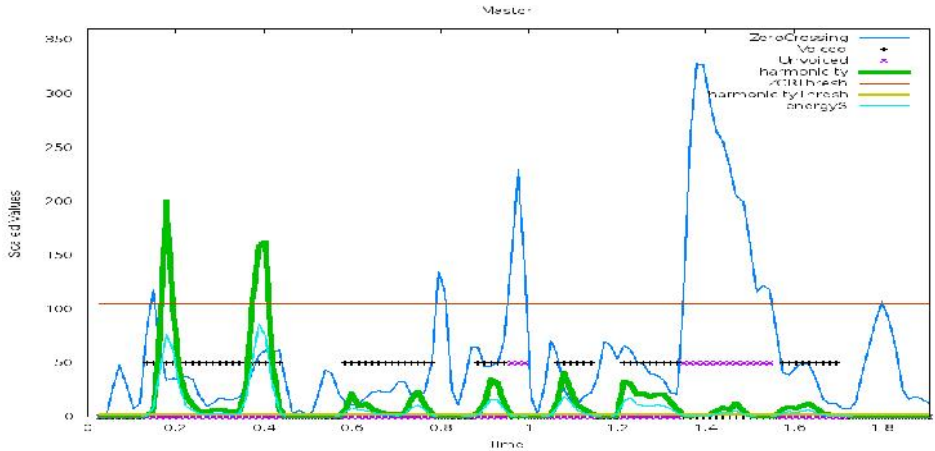

Fig. 2. Segmentation unit showing voiced and unvoiced parts.in a speech signal from EMO-DB

## 4. Feature Extraction

Extracting valuable features is another challenging task in the emotion recognition system. Mel frequency cepstral coefficients (MFCC) are one of the important features used in speech signal processing. Initially designed for speech recognition tasks they often give excellent performance in emotion detection tasks as well [13]. We therefore calculate the average, standard deviation and minima and maxima for 17 MFCCs. Next we extract Loudness features. The Loudness is a measure for how loud or how soft a sound is perceived-relatively independent of actual amplitude level of the signal. We apply bark filter banks [6] on the signal and then calculate loudness as described by Zwicker[4]. The center frequencies for the corresponding filters are calculated according to bark scale.

$$\Omega(k) = 6*\log\left[\frac{f(k)}{600} + \sqrt{\left(\frac{f(k)}{600}\right)^2 + 1}\right] \qquad (1)$$

fmin  f(k)  fmax

The shape of the filters is defined by equation:

$$Ck(w) = \begin{cases} 10^{1(\Omega-\Omega(k))+0.5}, \Omega \leq \Omega(k) - 0.5 \\ 1, \Omega(k) - 0.5 < \Omega < \Omega(k) + 0.5 \\ 10^{-2.5(\Omega-\Omega(k))-0.5}, \Omega \geq \Omega(k) + 0.5 \end{cases} \quad (2)$$

Ck(w) is a weight of the k filter at frequency w, $\Omega(k)$ is a center frequency of the filter, k = 1,2,3…..,K.

Next we calculate energy and pitch related features. To calculate the pitch we use cepstrum based pitch determination [7]. In general we observe that there is a peak in the cepstrum at the fundamental period of the input speech segment. The position of this peak should then correspond to the pitch period. We also use the short time energy features obtained during segmentation. For the energy and the loudness feature set we calculate the average, standard deviation and minima and maxima.

Next we calculate the above statistics on the first and second order derivatives of the contours in order to exploit temporal behavior at certain point of time.

Next we use the rhythm features used in linguistics including V, the standard deviation of vocalic internals (Vs); C, the standard deviation of consonantal internals (Cs); and %V, the relative duration of vocalic intervals within the total utterance etc. Our experiments seek to establish an analogy of vocalic and consonants intervals to the voiced and unvoiced parts respectively. Rhythm features namely mean voiced duration, mean unvoiced duration, mean pause duration, standard deviation of voiced, unvoiced, and silence intervals are calculated. VarcoV feature given by Eq.1 is then calculated for voiced, unvoiced and silence intervals.

$$VarcoV = std(V)*100/mean(V) \quad (3)$$

Similarly we calculate number of voiced intervals per sec and the rhythm feature nPVI for voiced, unvoiced and silence regions which is given by Eq.2 as:

$$nPVI = 100 * \sum_{k=1}^{m-1} \left| \frac{V(k) - V(k-1)}{(V(k) + V(k+1))/2} \right| /(m-1) \quad (4)$$

Next we calculate temporal features such as:
- duration of pause / duration of voiced+unvoiced
- duration of voiced/ duration of unvoiced
- duration of unvoiced / duration of unvoiced+ voiced
- duration of voiced / duration of unvoiced+ voiced
- duration of voiced/ pause(silence)
- duration of unvoiced / pause(silence)

In whole we get a set of 487 features. Table 1 shows a summary of extracted and calculated features and the number of features respectively.

Table 1. Extracted Features

| Feature Source | Number of Features |
|---|---|
| MFCC's voiced | 204 |
| MFCC's unvoiced | 204 |
| Loudness Voiced | 12 |
| Loudness Unvoiced | 12 |
| Pitch | 12 |
| Energy | 24 |
| Rhythm and Temporal | 19 |

## 5. Feature Selection

In order to determine the most promising features for our task individually, we applied an Information Gain Ratio (IGR) filter. We use WEKA toolkit .This entropy-based filter estimates the goodness of a single attribute by evaluating its information contribution (gain) with respect to the required mean information that leads statistically to a successful classification. The final ranking is obtained by using 10 fold cross validation. Table 2 shows the top 20 ranked features.

To select optimized number of features for classification we expand the feature space by including more and more features starting with the high ranks of the IGR output as shown in Figure 3. We observe that selecting 305 features gives the best recognition. As expected, the recognition rates rise by adding features until a global optimum is reached. Including more features beyond this optimal number causes a degradation of recognition rates again, due to adding irrelevant or even harmful information to the classification.

## 6. Classification

In our Classification experiments compare the classification results achieved by using different features separately as well as in combinations. We evaluate the performance of two different classifiers as shown in Table 3. The first classifier ANN is implemented following the multilayer perceptron architecture, using WEKA software. An artificial neural network (ANN), usually called "neural network" (NN), is a mathematical model or computational model that tries to simulate the structure and/or functional aspects of biological neural networks. It consists of an interconnected group of artificial neurons and processes information using a connectionist approach to computation [14]. After experimenting with different network parameters highest accuracy is found by using 200 neurons in hidden layer. The learning and momentum rate are left to the default setting of WEKA (0.3 and 0.2 respectively). The number of epochs is set to 500. Error backpropagation is used as a training algorithm. As a second classifier we choose a Support Vector Machine (SVM). John Platt's sequential minimal optimization algorithm is used for the optimizing and a Polynomial kernel of first order is used [11]. The value of cost parameter is kept to be 1. Now we do speaker dependent experiments using these classifiers.

Table 2. Information Gain Ranking

| Rank | Feature |
|---|---|
| 1 | mfcc_max_org_1_voiced |
| 2 | mfcc_mean_org_1_voiced |
| 3 | mfcc_mean_org_16_voiced |
| 4 | mfcc_mean_org_2_voiced |
| 5 | mfcc_min_org_1_unvoiced |
| 6 | mfcc_min_org_1_voiced |
| 7 | pitch_min_org_voiced |
| 8 | pitch_mean_org_voiced |
| 9 | mfcc_max_org_2_voiced |
| 10 | mfcc_mean_org_15_voiced |
| 11 | mfcc_mean_org_17_voiced |
| 12 | mfcc_max_org_15_voiced |
| 13 | mfcc_max_org_14_voiced |
| 14 | mfcc_mean_org_14_voiced |
| 15 | mfcc_max_org_13_voiced |
| 16 | mfcc_mean_org_5_unvoiced |
| 17 | mfcc_mean_org_6_unvoiced |
| 18 | rhythm_mean_voiced |
| 19 | mfcc_min_org_2_unvoiced |
| 20 | mfcc_max_org_10_voiced |

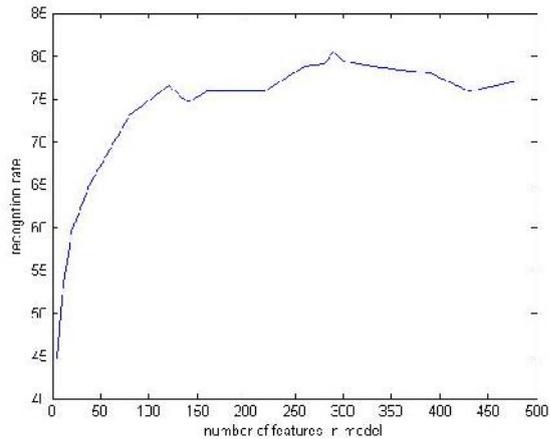

Fig.3. Recognition rate versus number of selected features from IGR ranking

We display results in form of confusion matrix and % recognition. After selecting the top 305 features by the selection algorithm described in previous section we are able to achieve recognition rate of 80.60 % on Berlin Emotion Database for 7 emotions classification problem. For the classification experiment the 10x10-fold stratified cross-validation method is employed over the data sets.

Table 3. Recognition rates for different classifiers

| Classifier | % Recognition |
|---|---|
| SVM | 80.27 % |
| ANN | 80.60% |
|  |  |

Table 4 and Table 5 show the confusion matrices for SVM and ANN. From Table 6 we interpret that Sadness is the best recognized emotion with accuracy of 95.2 %, 88.71 % and Happiness is worst recognized with accuracy of 66.2 %, 60.56 % for ANN and SVM classifier respectively. SVM classifier gives better accuracy for Boredom and Neutral Classes in comparison to ANN which gives better accuracy for Happiness, Sadness and Disgust Classes.

**Table 4**. Confusion matrix after feature selection (SVM Classifier)

| B | N | A | F | H | S | D | **EMOTION** |
|---|---|---|---|---|---|---|---|
| 66 | 9 | 0 | 0 | 0 | 2 | 2 | B (Boredom) |
| 5 | 67 | 0 | 4 | 2 | 1 | 0 | N (Neutral) |
| 0 | 1 | 106 | 2 | 14 | 0 | 3 | A (Anger) |
| 0 | 4 | 5 | 55 | 2 | 2 | 0 | F (Fear) |
| 2 | 2 | 18 | 5 | 43 | 0 | 1 | H (Happiness) |
| 2 | 3 | 0 | 1 | 0 | 55 | 1 | S (Sadness) |
| 3 | 2 | 2 | 2 | 1 | 2 | 34 | D (Disgust) |

**Table 5**. Confusion matrix after feature selection (MultilayerPerceptron)

| B | N | A | F | H | S | D | **EMOTION** |
|---|---|---|---|---|---|---|---|
| 61 | 10 | 0 | 0 | 0 | 4 | 4 | B (Boredom) |
| 10 | 62 | 1 | 2 | 1 | 2 | 1 | N (Neutral) |
| 0 | 0 | 106 | 3 | 14 | 0 | 3 | A (Anger) |
| 0 | 5 | 4 | 54 | 4 | 1 | 0 | F (Fear) |
| 0 | 2 | 16 | 5 | 47 | 0 | 1 | H (Happiness) |
| 1 | 0 | 0 | 1 | 0 | 59 | 1 | S (Sadness) |
| 1 | 0 | 2 | 3 | 1 | 0 | 39 | D (Disgust) |

Table 6. Class wise recognition rate for ANN and SVM classifier

| Emotion | SVM | ANN |
|---|---|---|
| Boredom | 83.54 % | 77.2 % |
| Neutral | 84.81 % | 78.48 % |
| Anger | 84.13 % | 84.13 % |
| Fear | 80.88 % | 79.42 % |
| Happiness | 60.56 % | 66.20 % |
| Sadness | 88.71 % | 95.2 % |
| Disgust | 73.91 % | 84.7 % |

Now we visualize the results obtained by different sets of features individually and in groups as shown in Table 7. We see that MFCC features alone are the best features giving a recognition rate of 71.93 %. Adding rhythm features to MFCC improves the accuracy to 74.02 % while the Rhythm features by themselves lead to only 34.6 %. However, as there are more than 10 times more MFCC features compared to the number of Rhythm features this result is expected to be influenced by quantity of features also. Loudness Features lead to only 44 % accuracy. If MFCC features are excluded then accuracy of 62.5 % is achieved. MFCC voiced (62 %) have better accuracy then MFCC unvoiced (49.3 %).

Table 7. Recognition rates for Different Sets of Features (SVM classifier)

| Features | Recognition Rate (in %) |
|---|---|
| MFCC only | 71.9 |
| Loudness + Rhythm | 52.6 |
| Loudness | 43.9 |
| Except MFCC | 62.5 |
| Rhythm only | 34.6 |
| MFCC+ Rhythm | 74.02 |
| MFCC unvoiced | 49.3 |
| MFCC voiced | 67.3 |

Next we break down the seven-class experiment into binary classification targets and get the results as shown in Table 8. Here we interpret that anger and happiness get confused very often because many of the features tend to show similar behavior for these two classes [13]. Best classification is achieved for happiness and sadness pair where accuracy of 100 percent is achieved.

Next we cluster the emotion classes into High Arousal (Happiness, Anger and Fear) and Low Arousal (Boredom, Sadness, Disgust, Neutral) and do the analysis. We interpret from the results in Table 9; Rhythm features all alone are able to give an accuracy of 74%, MFCC alone are able to give an accuracy of 89 %, MFCC and rhythm combined result in 92.4 % accuracy and an Overall accuracy of 94 % is achieved by selecting top 305 features by the same method as described in previous section.

**Table 8**. Recognition rates for binary classification targets (SVM classifier)

| Emotion | Recognition Rate (in %) |
|---|---|
| Anger vs. Happiness | 79.23 |
| Anger vs. Disgust | 95.56 |
| Anger vs. Fear | 92.04 |
| Anger vs. Boredom | 98.93 |
| Anger vs. Sadness | 99.41 |
| Anger vs. Neutral | 98.94 |
| Fear vs. Happiness | 92.5 |
| Fear vs. Disgust | 79.38 |
| Fear vs. Boredom | 98.52 |
| Fear vs. Neutral | 92 |
| Fear vs. Sadness | 95.83 |
| Disgust vs. Happiness | 92.92 |
| Disgust vs. Boredom | 92.37 |
| Disgust vs. Neutral | 96.63 |
| Disgust vs. Sadness | 95.098 |
| Happiness vs. Boredom | 99.307 |
| Happiness vs. Neutral | 93.75 |
| Happiness vs. Sadness | 100 |
| Boredom vs. Neutral | 92.42 |
| Boredom vs. Sadness | 96.79 |
| Sadness vs. Neutral | 98.94 |

**Table 9**. Recognition rates for High and Low arousal classes.

| Feature | Recognition rate |
|---|---|
| Rhythm | 74 % |
| MFCC | 89% |
| MFCC + Rhythm | 92.4% |

## 7. Conclusion

In this paper we utilize a new set of rhythm and temporal features and do several experimentations by using these features and other more conventional features. We are able to achieve a recognition rate of 80.60 % on Berlin emotion database with 7 different emotions in case of speaker dependent framework. We interpret from our results that adding rhythm and temporal features lead to increase in accuracy as compared to just using conventional features. In case of Binary classification targets we see that accuracy is above 90 percent for most of the pairs. Sadness and happiness pair is classified with 100 percent accuracy. We visualize that classifying the emotions into two sets (high and low arousal emotions) results in better accuracy. Investigation in [18] showed that the optimal feature set strongly depends on the emotions to be separated. Using a hierarchical classification strategy as suggested in [18], consisting of different classification stages distinguishing different classes, and using different feature sets will help in improving the accuracy. For this task experiments need to be done to interpret which emotions are better classified by traditional features and which emotions are classified well with the new set of rhythm and temporal features. A future work should incorporate an improved segmentation unit by employing fuzzy logic and neural networks so as to have a better classification of voiced and unvoiced segments and hence help fetch better features and lead to improvement in accuracy. Using a hybrid mixed classification ensemble can help in improvement of accuracy.